\documentclass{article}

\usepackage{PRIMEarxiv}

\usepackage[utf8]{inputenc} 
\usepackage[T1]{fontenc}    
\usepackage{url}            
\usepackage{booktabs}       
\usepackage{amsfonts}       
\usepackage{nicefrac}       
\usepackage{microtype}      
\usepackage{lipsum}
\usepackage{fancyhdr}       
\usepackage{graphicx}       
\graphicspath{{media/}}     
\usepackage{xcolor} 
\definecolor{lightblue}{rgb}{0.4, 0.6, 0.8} 
\usepackage{graphicx}
\usepackage{appendix} 

\usepackage[colorlinks=true, linkcolor=lightblue, citecolor=lightblue, urlcolor=lightblue]{hyperref} 
\usepackage{tikz}
\usetikzlibrary{positioning, arrows.meta, bending, shadows.blur}
\usepackage{placeins}
\usepackage{caption}
\usepackage{booktabs} 
\usepackage{array}    
\usepackage{amsmath}
\usepackage{booktabs} 
\usepackage{booktabs} 
\usepackage{array}    
\usepackage{multirow} 

\title{A short Survey: Exploring knowledge graph-based neural-symbolic system from application perspective}

\author{
  \begin{minipage}[t]{0.35\textwidth}
    \centering
    Shenzhe Zhu\thanks{First Author} \\
    University of Toronto \\
    Toronto, Canada \\
    \texttt{cho.zhu@mail.utoronto.ca}
  \end{minipage}
  \hfill
  \begin{minipage}[t]{0.35\textwidth}
    \centering
    Shengxiang Sun \\
    University of Toronto \\
    Toronto, Canada \\
    \texttt{owen.sun@mail.utoronto.ca}
  \end{minipage}
}

\begin{document}
\maketitle

\begin{abstract}
Advancements in Artificial Intelligence (AI) and deep neural networks have driven significant progress in vision and text processing. However, achieving human-like reasoning and interpretability in AI systems remains a substantial challenge. The Neural-Symbolic paradigm, which integrates neural networks with symbolic systems, presents a promising pathway toward more interpretable AI. Within this paradigm, Knowledge Graphs (KG) are crucial, offering a structured and dynamic method for representing knowledge through interconnected entities and relationships, typically as triples (subject, predicate, object). This paper explores recent advancements in neural-symbolic integration based on KG, examining how it supports integration in three categories: enhancing the reasoning and interpretability of neural networks with symbolic knowledge (Symbol for Neural), refining the completeness and accuracy of symbolic systems via neural network methodologies (Neural for Symbol), and facilitating their combined application in Hybrid Neural-Symbolic Integration. It highlights current trends and proposes future research directions in Neural-Symbolic AI.
\end{abstract}

\section{Introduction}
\label{Section 1}
With the rapid advancement of deep learning, particularly in deep neural networks (DNNs) within Artificial Intelligence (AI), we have observed the emergence of groundbreaking methods. These innovations have obtained significant achievements in fields such as vision and text processing. For instance, models like EfficientNet\cite{tan2019efficientnet}, ResNet\cite{zhang2022resnest}, and Vision Transformer\cite{dosovitskiy2020image} have demonstrated exceptional performance in tasks like image classification, target detection, and image segmentation. Similarly, the NLP domain has seen substantial strides with deep neural network-based pre-trained language models, such as GPT-4\cite{achiam2023gpt}, Llama 2\cite{touvron2023llama}, and BERT\cite{devlin2018bert}, setting new benchmarks in text comprehension and generation.

Despite these successes, the opacity of deep neural network models, often referred to as the "Black Box" problem\cite{ghorbani2019interpretation,castelvecchi2016can,linardatos2020explainable,molnar2020interpretable,lipton2018mythos, dayhoff2001artificial}, has obtained considerable attention. This problem arises when it becomes challenging to trace and elucidate the reasoning behind a model's decision-making. This opacity stems from the model's intricate internal structure, which involves millions of parameters that adjust automatically during training to optimally represent the input data, thereby complicating the understanding of the model's decision-making process. Addressing this issue is vital for fostering user trust, ensuring system fairness and security, and advancing AI technology's integrity.

Various scientists and researchers have proposed solutions to the black-box problem, extending into the realm of Explainable AI (XAI)\cite{arrieta2020explainable,ratti2022explainable}. Approaches like SHAP\cite{lundberg2017unified} and LIME\cite{ribeiro2016should} focus on feature attribution, aiming to clarify each input feature's contribution to the model's decision-making. Meanwhile, CAM\cite{zhou2016learning} and Grad-CAM\cite{selvaraju2017grad} utilize visualization techniques to demystify the model's internal mechanisms, aiding human understanding of how models process and interpret data. Additionally, SENN\cite{alvarez2018towards} adopts an Explanatory Embedded Modeling approach, enhancing explainability by integrating it into the model design phase, thereby creating a more transparent and logical model.

The field of model interpretability offers a vast array of research avenues, with some scientists recently try to explore the concept of integrating neural and symbolic systems to address the black-box problem\cite{daniele2022deep,diaz2022explainable,bennetot2019towards,ferreira2022looking,bennetot2022greybox,arabshahi2018combining,suddarth1988symbolic}. Yoshua Bengio, an ACM Turing Award laureate, highlighted in his 2019 NeurIPS presentation the necessity for deep learning to evolve from System 1 to System 2 thinking\cite{kahneman2011thinking}. System 1 refers to the intuitive, fast, and unconscious cognitive processes that current deep learning\cite{lecun2015deep} technologies excel in. In contrast, System 2 represents the logical, deliberate, and conscious cognitive processes, a hallmark of Symbolic artificial intelligence in the expert system stage\cite{smolensky1987connectionist,giarratano1998expert,patterson1990introduction}. This stage employs explicit symbols and rules to emulate human logical reasoning. This concept of transition underpins the concept of \textbf{neural-symbolic systems}, aiming to marry the pattern recognition prowess of deep learning models with the structured knowledge representation and logical reasoning capabilities of symbolic logic systems, thereby offering efficient abilities in learning and generalization, and clear logic. (As shown in \hyperref[table 1]{Table~\ref{table 1}} shows the strengths and drawbacks of the neural and symbolic systems)

Knowledge graph (KG)\cite{fensel2020introduction}, as an important member of symbolic logic, plays a crucial role in neural-symbolic integration. They are built on triples (subject, predicate, object) and form a graph structure that encapsulates real-world entities, concepts, and their interrelations. In neural symbolic systems, the KG not only serves as a repository of information but also acts as a bridge connecting symbolic logic and neural networks. It enriches the contextual information of the neural network and improves the decision-making and interpretability of the model, for example, in NLP tasks to help the network understand the relationship between words and entities. Meanwhile, symbolic reasoning relies on the logical rules and facts provided by the KG, which play an important role in both model training and reasoning. This idea of combining KG and neural symbolic integration enables us to build smarter, more reliable, and transparent AI systems.

The rest of this paper is organized as follows: Section \ref{Section2} introduces the major classifications of neural-symbolic systems. Section \ref{Section3} delves into several representative methods and models of neural-symbolic systems that incorporate KG. Section \ref{Section4} discusses the future trends in the field. Finally, Section \ref{Section5} concludes the paper.

In this paper, the term "neural system" primarily denotes deep neural networks\cite{lecun2015deep}. On the other hand, the "symbolic system" largely pertains to symbolic knowledge encapsulated within KG and related KG reasoning techniques. Additionally, in certain contexts, it may also encompass methods associated with symbolic reasoning.

\renewcommand{\arraystretch}{1.8}
\begin{table}
  \centering
  \begin{tabular}{>{\raggedright\arraybackslash}p{2.5cm} >{\raggedright\arraybackslash}p{3.5cm} >{\raggedright\arraybackslash}p{4.5cm} >{\raggedright\arraybackslash}p{4.5cm}}
    \toprule
    System & Core method & Advantage & Disadvantage \\
    \midrule
    Neural System & Data patterns learning & Strong representational capacity Handles complex patterns & Black box/Poor interpretability Relies on excessive data \\
    Symbolic System & Rule-based reasoning & Precise and logical \newline Highly interpretable & Less stable \newline Less flexible \\
    \bottomrule
  \end{tabular}
  \vspace{20pt}
  \caption{Comparison between neural system and symbolic system}
  \label{table 1}
\end{table}
\renewcommand{\arraystretch}{1} 

\section{Categorization of neural-symbolic systems}
\label{Section2}

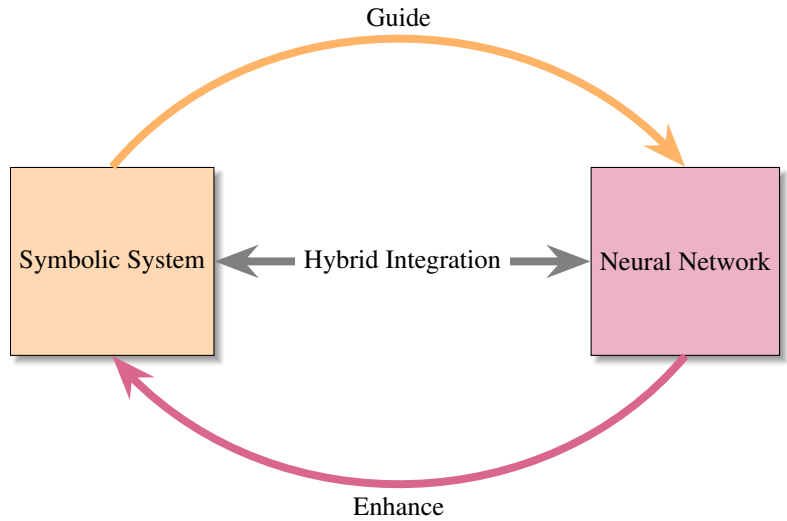
\begin{figure}[htb]
\centering
\begin{tikzpicture}[
    node distance=5 cm and 5 cm,
    mynode/.style={draw, rectangle, align=center, minimum height=2.5cm, minimum width=2.5cm, fill=#1, blur shadow={shadow blur steps=5}},
    myarrow/.style={-Stealth, bend left=50, draw=#1, line width=3pt},
    myarrowreverse/.style={-Stealth, bend left=50, draw=#1, line width=3pt},
    bidir/.style={Stealth-Stealth, draw=#1, line width=3pt}]
    
    \node[mynode=orange!30] (symbolic) {Symbolic System};
    \node[mynode=purple!30, right=of symbolic] (neural) {Neural Network};
    
    \draw[myarrow=orange!60] (symbolic.north) to node[midway, above] {Guide} (neural.north);
    \draw[myarrowreverse=purple!60] (neural.south) to node[midway, below] {Enhance} (symbolic.south);
    \draw[bidir=gray] (symbolic.east) -- (neural.west) node[midway, fill=white] {Hybrid Integration};

\end{tikzpicture}
\caption{This diagram depicts three neural-symbolic system interactions: the orange curve for "Symbol for Neural", the pink for "Neural for Symbol", and the grey bidirectional line for "Hybrid neural-symbolic integration", highlighting their distinct collaborative dynamics.}
\label{figure 1}
\end{figure}

Delving into the categorization of neural-symbolic systems unveils three primary interaction models\cite{sun2013connectionist,mcgarry1999hybrid}: Symbol for Neural, Neural for Symbol, and Hybrid Neural-Symbolic Integration. Each category represents a distinct approach to integrating neural and symbolic components (see \hyperref[figure 1]{Figure~\ref{figure 1}}). In this section, we will explore the definitions, and frameworks within these three taxonomies and how KG can be integrated into these systems.

\subsection{Neural for symbol}
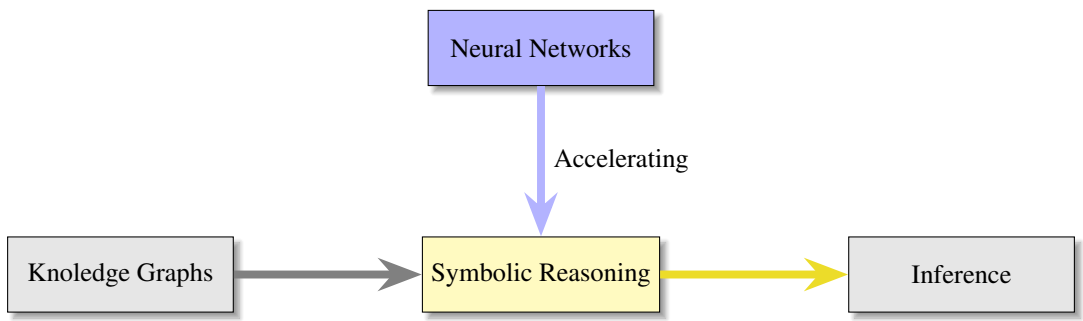
\begin{figure}[htb]
\centering
\begin{tikzpicture}[
    node distance=2 cm and 2.5 cm,
    mynode/.style={draw, rectangle, align=center, minimum height=1cm, minimum width=3cm, fill=#1, blur shadow={shadow blur steps=5}},
    myarrow/.style={-Stealth, draw=gray, line width=3pt},
    myarroworange/.style={-Stealth, draw=yellow!90!black, line width=3pt},
    myarrowpurple/.style={-Stealth, draw=blue!30, line width=3pt}
]

\node[mynode=gray!20] (input) {Knoledge Graphs};
\node[mynode=yellow!30, right=of input] (symbolic) {Symbolic Reasoning};
\node[mynode=blue!30, above=of symbolic] (neural) {Neural Networks};
\node[mynode=gray!20, right=of symbolic] (output) {Inference};

\draw[myarrow] (input) -- (symbolic);
\draw[myarrowpurple] (neural) -- node[midway, right] {Accelerating} (symbolic);
\draw[myarroworange] (symbolic) -- (output);

\end{tikzpicture}
\caption{Neural for symbol}
\label{figure 3}
\end{figure}

"Neural for Symbol", also known as "Learning for Reasoning", focuses on utilizing the learning capabilities of neural networks for the enhancement and problem solving of traditional symbolic reasoning. In this paradigm, neural networks usually enhance symbolic systems by acceleration \cite{kim2020integration,das2016chains,xiong2017deeppath,meilicke2020reinforced,chen2018variational,lao2011random,neelakantan2015compositional,lin2018multi,zhu2021neural,teru2020inductive,cohen2016tensorlog,yang2017differentiable,yang2019learn,qu2020rnnlogic,zhang2020efficient,qu2019probabilistic,wang2019knowledge,wang2019kgat,zhang2018variational, sun2018open, sun2019pullnet,yasunaga2021qa,liu2021contextualized,li2021learning}.Typically, "acceleration" refers to the use of neural networks to improve the speed and efficiency of symbolic systems in knowledge reasoning and processing complex data. For instance, Neural networks can optimize the search path\cite{zhang2021neural} in KG reasoning by leveraging their advanced analytical capabilities. \hyperref[figure 3]{Figure~\ref{figure 3}} illustrates the architecture of this integrated approach.

\subsection{Symbol for neural}
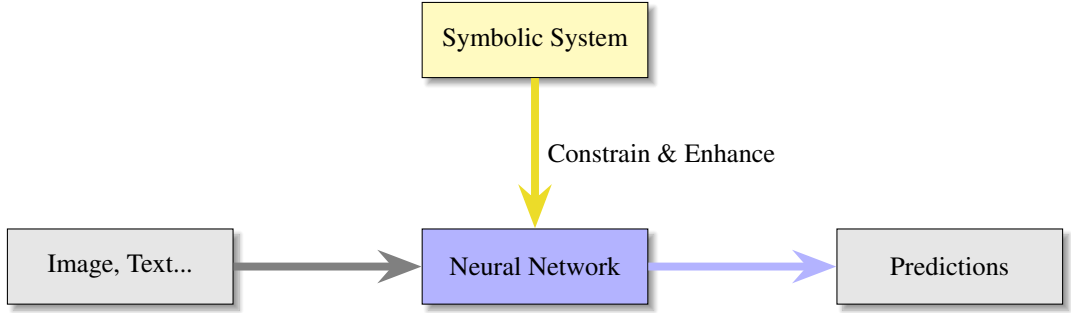
\begin{figure}[htb]
\centering
\begin{tikzpicture}[
    node distance=2 cm and 2.5 cm,
    mynode/.style={draw, rectangle, align=center, minimum height=1cm, minimum width=3cm, fill=#1, blur shadow={shadow blur steps=5}},
    myarrow/.style={-Stealth, draw=gray, line width=3pt},
    myarroworange/.style={-Stealth, draw=yellow!90!black, line width=3pt},
    myarrowpurple/.style={-Stealth, draw=blue!30, line width=3pt}
]

\node[mynode=gray!20] (input) {Image, Text...};
\node[mynode=blue!30, right=of input] (neural) {Neural Network};
\node[mynode=yellow!30, above=of neural] (symbolic) {Symbolic System};
\node[mynode=gray!20, right=of neural] (output) {Predictions};

\draw[myarrow] (input) -- (neural);
\draw[myarroworange] (symbolic) -- node[midway, right] {Constrain \& Enhance} (neural);
\draw[myarrowpurple] (neural) -- (output);

\end{tikzpicture}
\caption{Symbol for neural}
\label{figure 2}
\end{figure}

In this context, "Symbol for Neural"\cite{guo-etal-2016-jointly, guo2018knowledge, zhang2019iteratively, donadello2017logic, chen2013learning, serafini2016logic, xu2018semantic,cranmer2020discovering, hinton2012improving,wang2018ripplenet,zhang2020cazsl,hooshyar2024augmenting,daniele2022deep,hersche2023neuro,xie2019embedding,xie2021embedding,hu2016harnessing,padmanabhan2023lsfsl,diligenti2017semantic,kampffmeyer2019rethinking,chen2020knowledge,yao2023se,wang2019knowledge,wang2019kgat,zhang2016collaborative,sun2019pullnet,sun2018open,saxena2020improving,
lee2018multi,chen2020knowledge,kampffmeyer2019rethinking,
liu2020k,zhang2019ernie,lee2018multi,wang2018zero,kampffmeyer2019rethinking,yao2020graph,
liu2020k,zhang2019ernie}, also known as "Reasoning for Learning", leverages symbolic systems like KG to furnish a prior knowledge and a logical framework, thereby guiding and shaping neural networks' learning processes. Symbolic knowledge encoded in KG provides a rich source of structured information that enables neural networks to enhance their interpretability and decision-making capabilities. Moreover, KG serves not merely as passive repositories but as active participants, infusing neural networks with domain-specific rules and facts to bolster their learning efficiency. For instance, in developing algorithms for a recommendation system on an online education platform, a KG can categorize courses by content, difficulty, and progression, directing the neural network to tailor learning paths for users, thus optimizing learning outcomes. \hyperref[figure 2]{Figure~\ref{figure 2}} illustrates the architecture and rationale of this integrated approach.

\subsection{Hybrid neural-symbolic integration}
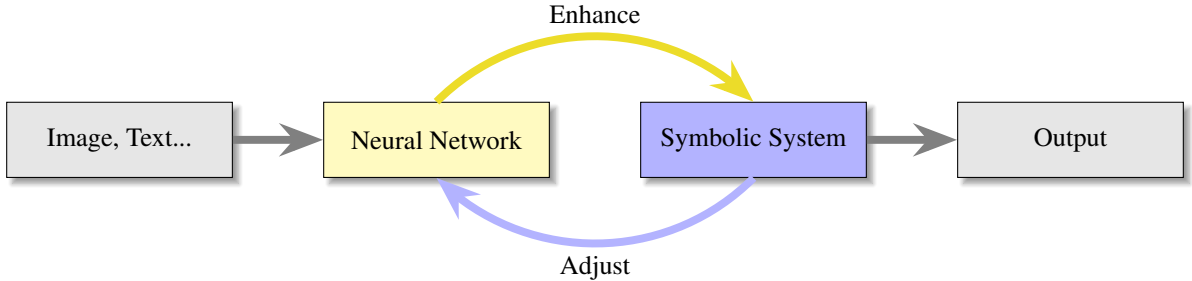
\begin{figure}[htb]
\centering
\begin{tikzpicture}[
    node distance=1 cm and 1.2 cm,
    mynode/.style={draw, rectangle, align=center, minimum height=1cm, minimum width=3cm, fill=#1, blur shadow={shadow blur steps=5}},
    myarrow/.style={-Stealth, draw=gray, line width=3pt},
    myarroworange/.style={-Stealth, draw=yellow!90!black, line width=3pt},
    myarrowpurple/.style={-Stealth, draw=blue!30, line width=3pt}
]

\node[mynode=gray!20] (input) {Image, Text...};
\node[mynode=yellow!30, right=of input] (neural) {Neural Network};
\node[mynode=blue!30, right=of neural] (symbolic) {Symbolic System};
\node[mynode=gray!20, right=of symbolic] (output) {Output};

\draw[myarrow] (input) -- (neural);
\draw[myarroworange] (neural.north) to[out=45,in=135] node[midway, above] {Enhance} (symbolic.north);
\draw[myarrowpurple] (symbolic.south) to[out=-135,in=-45] node[midway, below] {Adjust} (neural.south);
\draw[myarrow] (symbolic) -- (output);

\end{tikzpicture}

\caption{Hybrid neural-symbolic integration}
\label{figure 5}
\end{figure}

"Hybrid neural-symbolic integration"\cite{manhaeve2018deepproblog,zhou2019abductive,dai2019bridging,huang2021fast,johnson2013hybrid,cai2021abductive,tian2022weakly,harnad2001grounding,romaniuk1993sc,goertzel2009opencog,tsamoura2021neural,pisano2020neuro,cambria2022senticnet, ding2019cognitive,
ke2021jointgt,
zhang2023iterative,
zhao2023knowledge} shows a more dynamic way of interaction. In this approach, neural networks and symbolic reasoning complement each other without being subordinate, working together to enhance the comprehension and inference capabilities of the AI system. In this system, neural networks first process input data (e.g., images, text, etc.), extracting features and converting them into intermediate representations. These representations are then passed to a symbolic system, which utilizes this data for logical reasoning and possibly combines it with existing KG to make decisions or generate new knowledge. The results of the reasoning are not only used for direct decision-making output but are also fed back to the neural network to guide its further learning and parameter tuning, thus optimizing the performance of the overall system. Through this iterative feedback mechanism, the hybrid system can continuously optimize itself, and its components, the neural network and the symbolic system, can learn and adapt from each other's processing results. This collaborative process grants the hybrid system resilience and adaptability, allowing it to efficiently manage complex tasks with accuracy and interpretability. \hyperref[figure 5]{Figure~\ref{figure 5}} illustrates the architecture of this integrated approach.

\section{Methods based on knowledge graph}
\label{Section3}
In this section, we take an in-depth look at the three taxonomies of neural symbols introduced in Section \ref{Section2}, focusing on specific methods for applications. We aim to describe these representative approaches for combining neural symbols with KG techniques, demonstrating progress in several directions. In addition, \hyperref[table 2]{Table~\ref{table 2}} summarizes the relevant features of these representative methods.

\renewcommand{\arraystretch}{1.8} 

\begin{table}[htb]
  \centering
  \begin{tabular}{>{\raggedright\arraybackslash}p{3cm} >{\raggedright\arraybackslash}p{4cm} >{\centering\arraybackslash}m{4.5cm}}
    \toprule
    Model & Application & Categories \\
    \midrule
    KGCN\cite{wang2019knowledge} & \multirow{4}{*}{Recommender systems} & \multirow{8}{4.5cm}{\centering Neural for symbol} \\
    KGAT\cite{wang2019kgat} & & \\
    CGAT\cite{liu2021contextualized} & & \\
    HRAN\cite{li2021learning} & & \\
    \cline{1-2}
    VRN\cite{zhang2018variational} & \multirow{4}{*}{Q\&A systems} & \\
    GRAFT-Net\cite{sun2018open} & & \\
    PullNet\cite{sun2019pullnet} & & \\
    QA-GNN\cite{yasunaga2021qa} & & \\
    \cmidrule{1-3}
    SEKG-ZSL\cite{wang2018zero} & \multirow{4}{*}{Zero-shot and few-shot learning} & \multirow{7}{4.5cm}{\centering Symbol for neural} \\
    ML-ZSL\cite{lee2018multi} & & \\
    DGP\cite{kampffmeyer2019rethinking} & & \\
    GFL\cite{yao2020graph} & & \\
    \cline{1-2}
    K-BERT\cite{liu2020k} & \multirow{2}{*}{Knowledge-enhanced LMs} & \\
    KnowBERT\cite{peters2019knowledge} & & \\
    \cmidrule{1-3}
    CogQA\cite{ding2019cognitive} & \multirow{1}{*}{Q\&A systems} & \multirow{4}{4.5cm}{\centering Hybrid integration} \\
    \cline{1-2}
    JointGT\cite{ke2021jointgt} & \multirow{1}{*}{KG-to-text}& \\
    \cline{1-2}
    HGNN-EA\cite{zhang2023iterative} & \multirow{1}{*}{Entity alignment}& \\
    \cline{1-2}
    KIG\cite{zhao2023knowledge}  & \multirow{1}{*}{Sentiment identification}& \\

    \bottomrule
  \end{tabular}
  \vspace{20pt}
  \caption{Overview of Models by Application and Category}
  \label{table 2}
\end{table}

\renewcommand{\arraystretch}{1} 

\subsection{Neural for symbol}
Deep learning plays a pivotal role in enhancing KG-related applications. In the field of neural for symbol systems, networks significantly accelerate the efficiency of KG's symbolic reasoning. This integration can be categorized into two main categories based on their specific applications: recommender systems enhanced by KG\cite{wang2019knowledge,wang2019kgat,liu2021contextualized,li2021learning}, Q\&A systems enhanced by KG insights\cite{zhang2018variational, sun2018open,sun2019pullnet,yasunaga2021qa} These categories reflect the different ways in which neural networks promote the development of KG in various uses, enabling the building of faster and high-efficient computational models.

\subsubsection{KG-based recommender systems}
In the field of exploring KG-enhanced recommender systems, research has been divided into three main directions: path-based approaches\cite{yu2014personalized,zhao2017meta,shi2018heterogeneous}, embedding-based\cite{wang2018dkn,zhang2016collaborative,wang2018shine} approaches and propagation-based approaches\cite{wang2019kgat,wang2019knowledge} that we will introduce below.

Traditional KG recommendation systems like NFM\cite{he2017neural}, Wide\&Deep\cite{cheng2016wide}, and xDeepFM\cite{lian2018xdeepfm} struggle with non-linear relationships and high-order interactions. The \textbf{Knowledge Graph Convolutional Network (KGCN)}\cite{wang2019knowledge}, leveraging the Graph Convolutional Network (GCN)\cite{schlichtkrull2018modeling,kipf2016semi} framework, addresses these challenges by capturing multi-hop relationships between entities through stacked graph convolutional layers and a weighted neighbor aggregation mechanism. Also, this methodology enhances the understanding of complex entity relationships and, by utilizing GCN's neighborhood sampling and parallel computation, significantly improves computational efficiency.

\begin{figure}[htb] 
\centering 
\includegraphics[width=0.3\textwidth]{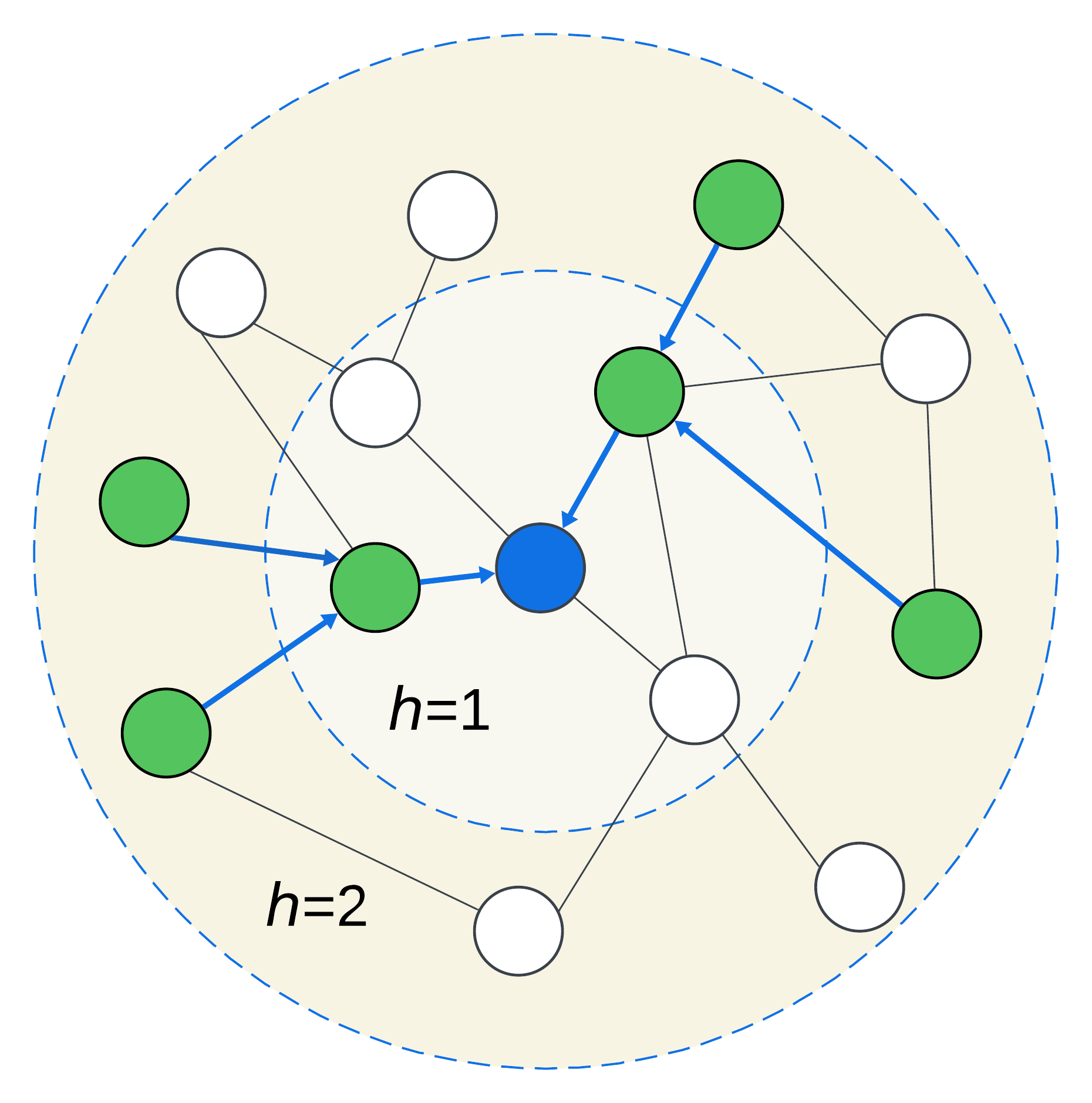} 
\caption{A two-layer receptive field (green entities) of the blue entity in a KG} 
\label{figure 6}
\end{figure}

Neighborhood aggregation in KGCN updates node representations by utilizing neighbor features and capturing high-order user interests within the KG. It involves calculating and normalizing user relevance scores to weight neighboring entities, allowing the model to consider both direct and multi-hop neighbors, thus broadening its influence field and capturing complex entity dependencies. This approach is depicted in a two-layer receptive field illustration(see \hyperref[figure 6]{Figure~\ref{figure 6}}), showcasing the multi-hop neighbor consideration. By incrementally constructing the network and employing negative sampling and gradient descent, KGCN refines entity representations. Additionally, domain sampling(uniformly sampling a fixed-sized set from each entity's neighbors)in each layer speeds up neighborhood information aggregation and propagation, enhancing the model's ability to quickly learn intricate patterns of entity interactions and relationships with fewer computational resources.

Similarly, \textbf{Knowledge Graph Attention Network (KGAT)}\cite{wang2019kgat} works on complex network situations between items due to shared attributes or characteristics that traditional methods ignore. By leveraging the adaptive focusing of relevant node property of graph attention network\cite{velickovic2017graph}, KGAT deeply mines the high-order relationships in the KG, significantly enhancing the model's understanding of the interrelationships between items, thereby improving the accuracy and relevance of recommendations.

\begin{figure}[htb] 
\centering 
\includegraphics[width=0.4\textwidth]{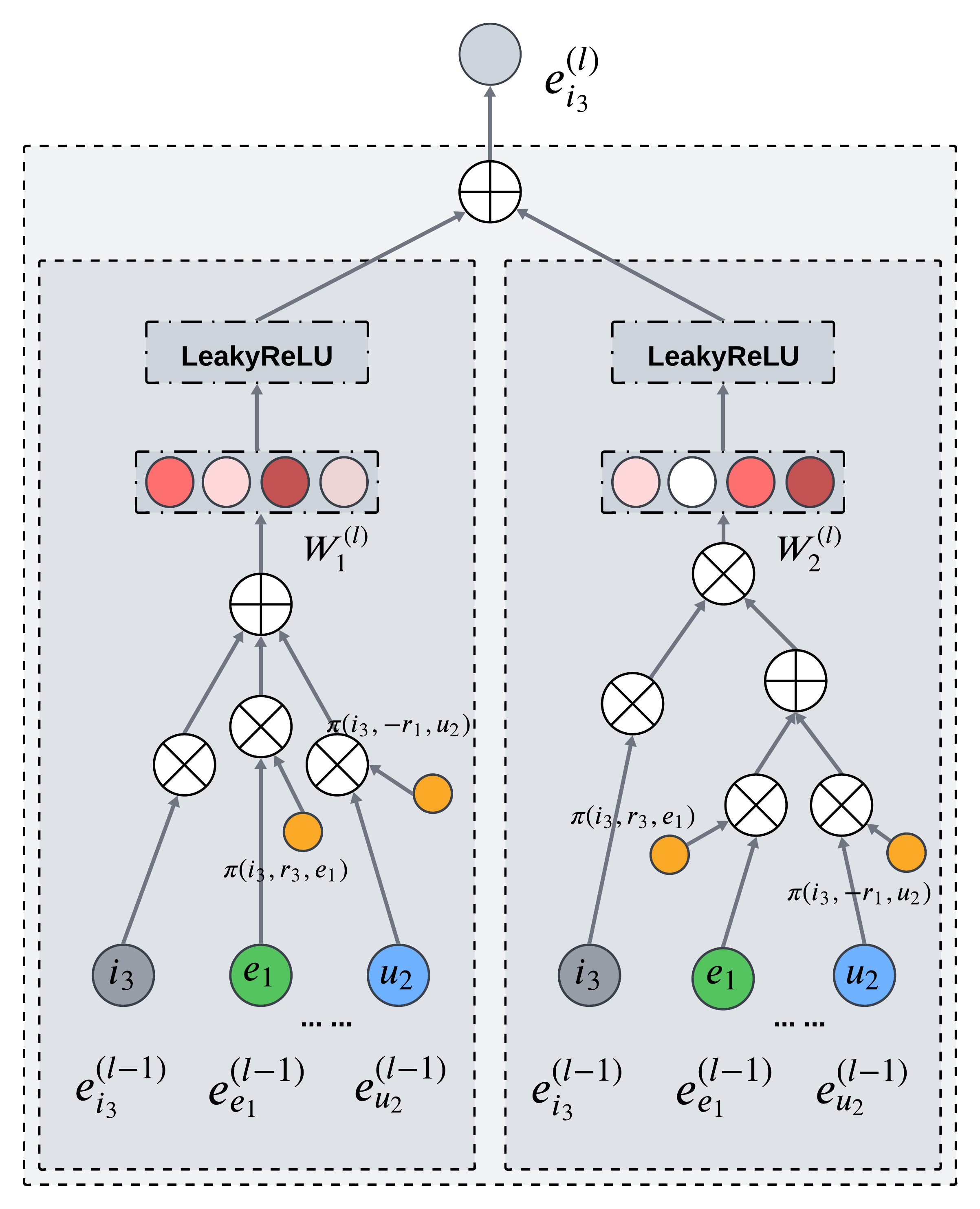} 
\caption{Attentive embedding propagation layer of KGAT} 
\label{figure 7}
\end{figure}

In addition to the basic collaborative KG embedding and prediction layers, the structure of KGAT specifically introduces an attention embedding propagation layer(see \hyperref[figure 7]{Figure~\ref{figure 7}}). This layer efficiently captures high-order relationships by integrating recursive embedding propagation and attention mechanisms. The inclusion of an attention mechanism allows the model to differentiate the importance of different neighboring nodes (see Attention Coefficient $\pi(h,r,t)$ below). Such a mechanism allows the neural network to focus more on more important nodes, thus improving the efficiency and accuracy of the recommender system. Firstly, recursive embedding propagation allows the model to gradually update the embedding representation of a node by considering the embedding information of the node and its neighbors, implemented through the equation \ref{eq 1}, where $e_{\mathcal{N}_h}$ represents the aggregated embedding of node h's first-order neighborhood, $e_t$ is the embedding of the neighbor node t, and $\pi(h,r,t)$ is the contribution weight of node t to h, reflecting the strength of the relationship between nodes.
\begin{equation}
\label{eq 1}
    e_{\mathcal{N}_h} = \sum_{(h,r,t) \in \mathcal{N}_h} \pi(h,r,t) e_t
\end{equation}
Subsequently, for the aforementioned attention coefficient \(\pi(h, r, t)\), we derive it with normalization using Equation \ref{eq 2}:
\begin{equation}
\label{eq 2}
    \pi(h, r, t) = \frac{\exp\left( (W_r e_t)^\top \tanh(W_r e_h + e_r) \right)}{\sum_{(h, r', t') \in N_h} \exp\left( (W_{r'} e_{t'})^\top \tanh(W_{r'} e_h + e_{r'}) \right)}
\end{equation}
where \(W_r\) is the transformation matrix for relation \(r\), \(e_h\) and \(e_t\) represent the embedding vectors for the head and tail entities, respectively, and \(e_r\) is the embedding vector for relation \(r\). Here, \(\tanh\) is employed as the activation function to help the model capture complex nonlinear relationships between entities while maintaining the output within a stable range of values. After these steps, we aggregate the information to update the entity's representation, and the model could recursively extend this information aggregation to more distant neighbors through high-order propagation, allowing each node's embedding to capture a broader context and accelerating the information transfer process throughout the KG.

In recent years, along with the development of graph attention network technologies such as KGAT, we have witnessed the rise of technologies such as \textbf{Contextualized Graph Attention Network(CGAT)}\cite{liu2021contextualized} and \textbf{Heterogeneous Relation Attention
Networks(HRAN)}\cite{li2021learning}, which further extend the application of graph attention networks in the field of KG reasoning. CGAT greatly enhances the performance of recommender systems by fusing local and non-local contextual information in the project KG. It utilizes a user-specific graph attention mechanism to aggregate neighborhood information in the KG, while taking into account the user's personalized preferences, enabling the model to provide customized recommendation services based on different users' attention to neighboring entities. HRAN, on the other hand, is designed for Heterogeneous KG Embedding (KGE), which operates at multiple semantic levels and hierarchically aggregates neighborhood features, while fully taking into account the information diversity of the KG. By introducing an innovative framework, HRAN uses an attention mechanism to determine the importance of different relational paths, enabling selective aggregation of information features.

\subsubsection{KG-based Q\&A systems}
KGs play a central role in the construction of contemporary Q\&A systems. By integrating and utilizing KGs, Q\&A systems can go beyond simple fact retrieval to achieve advanced question processing that requires in-depth semantic understanding and reasoning, while at the same time, we can leverage the learning capability of neural symbols to enhance the inference speed of KGs, thus significantly improving the response quality and user interaction experience.

Past KG-driven Q\&A systems\cite{miller2016key,weston2014memory,li2015gated} faced two major challenges: first, it is difficult to utilize the structural information of the KG for complex multi-hop logical reasoning; second, it is difficult to accurately locate the topic entities mentioned in the question in the presence of various noises. To address these problems, Y Zhang et al. introduce the \textbf{Variational Reasoning Network (VRN)} \cite{zhang2018variational}, a model grounded in a probabilistic modeling framework. It leverages a deep learning architecture that resembles a propagation mechanism, specifically tailored for logic reasoning over KG. Also, this model incorporates the REINFORCE algorithm, complemented by a variance reduction technique, to optimize its performance and reliability in inference tasks. 

Specifically, VRN uses the probabilistic framework to handle uncertainty, consisting of two modules. The first one is the " Module for topic entity recognition ". In this module, we use a neural network model 
\( f_{\text{ent}}(\cdot): q \mapsto \mathbb{R}^d\) that maps the problem to a high-dimensional vector space, thus capturing the problem context to identify and parse unique topic entities, rather than relying solely on pre-annotation or exact matching. The next one is the "Module for logic reasoning over KG". In this module, VRN uses the inference graph embedding architecture to solve the multi-hop problem and simplifies multi-step traversal in large KGs. It creates subgraphs that encapsulate all possible paths by performing topological ordering within a maximum number of hops of subject entities. It then learns the nonlinear embedding of these paths in the vector space. From there, it performs efficient reasoning on complex queries without the need for exhaustive graph traversal. Overall, this architecture avoids blind searches in large KG by combining probabilistic models with neural networks to predict potential inference paths.

\begin{figure}[htb] 
\centering 
\includegraphics[width=1\textwidth]{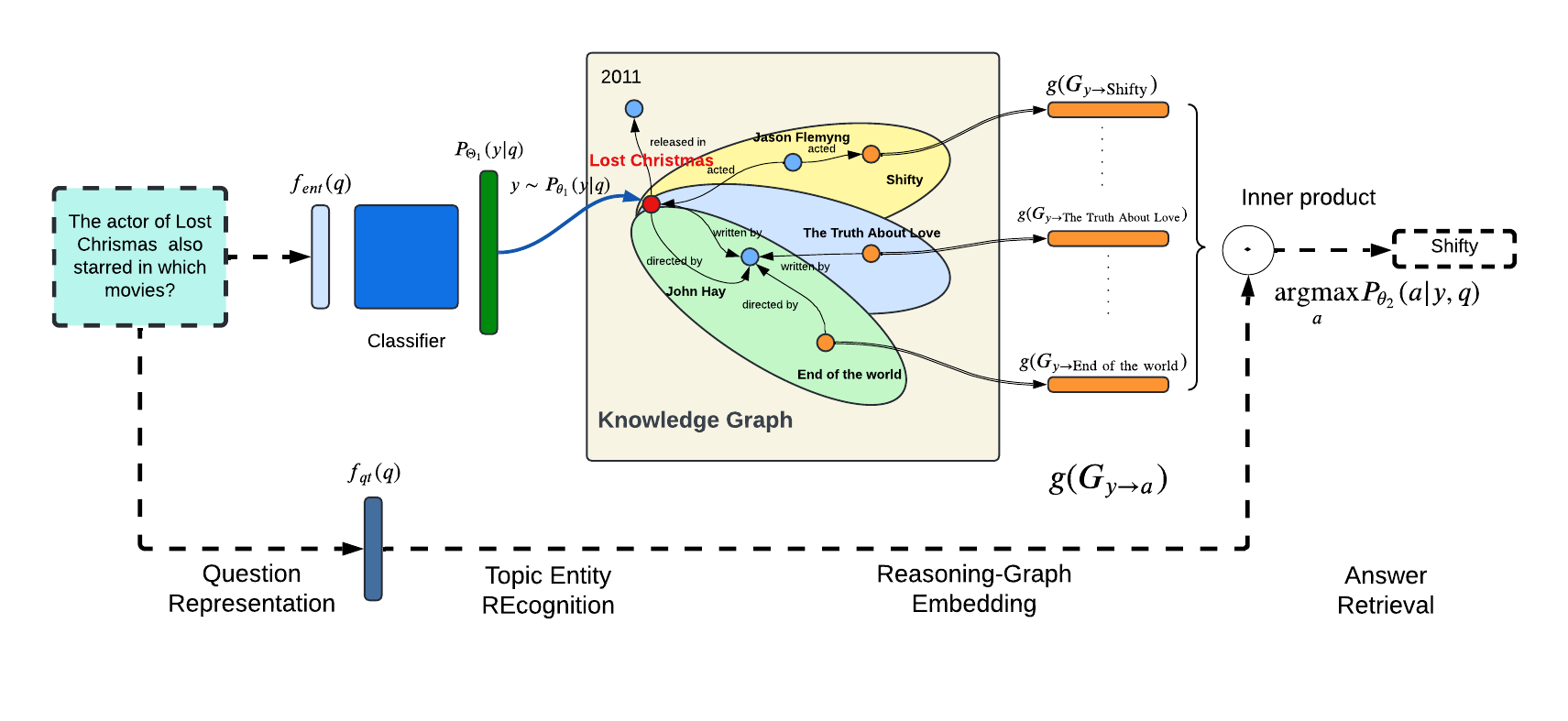} 
\caption{End-to-end architecture of the variational reasoning network (VRN) for question-answering with KG} 
\label{figure 8}
\end{figure}

Q\&A models are easy to encounter limitations due to their reliance on data extraction from a single source\cite{wiese2017neural, dhingra2018simple,min2013distant}. The \textbf{Graph of Facts Relationships and Texts Network (GRAFT-Net)}\cite{sun2018open} released by H Sun et al. transcends these limitations by fusing two different sources of information: KG and textual data, thus improving Q\&A performance. GRAFT-Net innovatively employs GCN to analyze heterogeneous graphs - a complex structure that combines both sources, encompassing learning the representation of different node types (entities and sentences) and their interconnections, thus allowing the neural network to deepen its understanding and reasoning about the interrelationships present in the composite graph. \hyperref[figure 8]{Figure~\ref{figure 8}} refers to the architecture of VRN.

Updating the heterogeneous graph constitutes a crucial phase in GRAFT-Net's training, where the refinement of entity nodes is pivotal. This update process employs a feed-forward network. Specifically, the update for an entity \(v\) is computed by amalgamating its previous state \(h_v^{(l-1)}\), the question's representation \(h_q^{(l-1)}\), and the collective states from neighboring entities \(N_r(v)\). These are weighted by attention coefficients \(\alpha_{r}^{v'}\) and modified by relation-specific transformations \(\psi_r\)\cite{schlichtkrull2018modeling}. Additionally, the update incorporates aggregated states from the entity's mentions across the documents \(M(v)\), ensuring a comprehensive update mechanism as delineated in equation \ref{eq 3}:
\begin{equation}
\label{eq 3}
h_v^{(l)} = \text{FFN}\left( \left[ \begin{array}{c} h_v^{(l-1)} \\ h_q^{(l-1)} \\ \sum_{r} \sum_{v' \in N_r(v)} \alpha_{r}^{v'} \psi_r(h_{v'}^{(l-1)}) \\ \sum_{(d,p) \in M(v)} H_{d,p}^{(l-1)} \end{array} \right] \right)
\end{equation}
Following the update rules, we continue into two integral techniques employed in this study: the attention mechanism and the directed propagation technique. These methods are crucial for sharpening the model's focus on graph regions pivotal to the query at hand. Through the attention mechanism, where weights:
\begin{equation}
    \alpha_{rv} = \text{softmax}\left(x_r^T h_q^{(l-1)}\right) = \frac{\exp(x_r^T h_q^{(l-1)})}{\sum_{k \in N_r(v)} \exp(x_k^T h_q^{(l-1)})}
\end{equation}
reflect the congruence between relation vectors and the question representation, the model's information flow is channeled along edges deemed relevant. Simultaneously, the directed propagation technique, inspired by personalized PageRank\cite{brin1998anatomy}, ensures targeted dissemination of embeddings from question-related seed nodes across pertinent graph paths, thereby maintaining the model's concentrated attention on essential areas of the graph.

Following the development of Graft-Net, the introduction of \textbf{PullNet}\cite{sun2019pullnet} by H. Sun's team marks a significant advancement in the field of Q\&A system. PullNet improves upon Graft-Net by introducing an iterative retrieval process. Instead of using fixed heuristics to construct a question-specific subgraph as in Graft-Net, PullNet employs a GCN to dynamically identify and expand relevant nodes in the subgraph. This method allows PullNet to efficiently gather pertinent information from both KBs and textual data, ensuring that the retrieved subgraph is comprehensive, and containing all the necessary information. Specifically, this iterative process begins with a basic subgraph containing only entities directly related to the problem. Subsequently, through a series of iterative steps, the system gradually expands this subgraph. In each iteration, PullNet uses GCN to assess the importance of each node in the subgraph and selects the k nodes most likely to help answer the question for expansion. This selection is based on the probability scores of each node, which are computed by the classification operation. For each selected node, PullNet performs a "pull" operation to retrieve new information related to these nodes, including facts from the knowledge base and documents from the corpus. The retrieved new information is then added to the subgraph, including not only the new nodes but also the edges connecting them.

With the rapid growth of pre-trained language models, many researchers have explored integrating them with KG for QA systems like \textbf{QA-GNN}\cite{yasunaga2021qa}. This model merges pre-trained language models and graph neural networks(GNN)\cite{scarselli2008graph} to enhance understanding of question-answering contexts and utilize KG effectively. Initially, QA-GNN interprets the context using a pre-trained model, then forms a joint graph with the KG. It assesses node relevance within the KG to the QA context using a scoring mechanism. Finally, the model applies GNNs and relevance scores for reasoning on the joint graph to predict answers.

The core innovations of QA-GNN are twofold. First, relevance scoring assesses the importance of each KG node in relation to the QA context(see \hyperref[figure 9]{Figure~\ref{figure 9}}). This scoring informs the attention mechanism within the graph neural network, enhancing node representation updates. For example, with a QA context node \(z\) and a KG node \(v\), the relevance score \(\rho_v\) is computed using
\begin{equation}
 \rho_v = f_{\text{head}}(f_{\text{enc}}([text(z); text(v)]))
\end{equation}
, where \(f_{\text{enc}}\) is the encoder extracting text features and \(f_{\text{head}}\) predicts the relevance score of node \(v\) to context \(z\). Second is joint reasoning, where after forming the joint graph, the model uses a graph neural network with an attention mechanism to update node representations, iteratively updating the representations of the context and KG, thereby enabling reasoning. In this process, the graph attention network (GAT)\cite{velickovic2017graph} dynamically adjusts the weights of information transfer between nodes so that each node can update its own information based on its relevance score with neighboring nodes.

\begin{figure}[htb] 
\centering 
\includegraphics[width=1\textwidth]{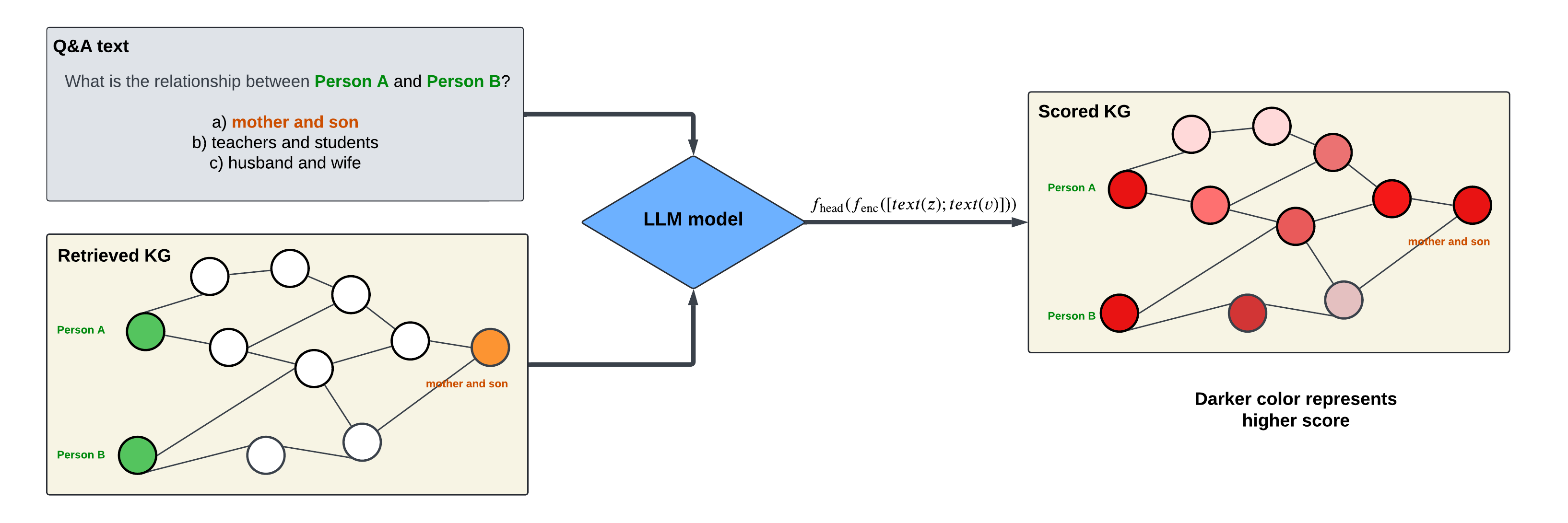} 
\caption{Architecture of the QA-GNN for question-answering with KG} 
\label{figure 9}
\end{figure}

\textbf{Conclusion:} Recommender systems and Q\&A systems, as the core applications combining KGs and neural networks, demonstrate a consensus on the choice of network models to utilize. First, these systems commonly adopt GNNs as their foundation, as they can resolve complex entity relationships and topological information directly on graph structures, providing a clear framework for efficiently processing KG data. For example, GCN-based\cite{kipf2016semi} models such as KGCN\cite{wang2019knowledge} perform convolutional operations on the graph to deeply learn the interactions between nodes and highlight the learning of local connections. And GAT-based\cite{velickovic2017graph} models, such as KGAT\cite{wang2019kgat} and QA-GNN\cite{yasunaga2021qa}, introduce the attention mechanism to improve the judgment of the importance of neighbors and achieve a detailed node representation. Second, multi-hop inference is a core technique in neural network-based KG reasoning, which plays an indispensable role in almost all KG applications. By performing multi-step logical reasoning in the KG, this approach can explore and reveal complex, multi-level relationships between data. In recommender systems, multi-hop reasoning helps the system to deeply understand the user's preferences and needs to achieve personalized recommendations, while in Q\&A systems, it enables the system to handle more complex queries and provide more accurate and insightful answers. Overall, all these models utilize the information and hierarchical relationships of graph structures to learn the representation of complex entities and relationships. Whether it is local structure capture in GCN, differential attention in GAT, or deep mining in multi-hop reasoning, they all demonstrate their ability to deal with complex and deep information, demonstrating their efficacy in intelligent applications.

\subsection{Symbol for neural}
The reasoning ability of the KG can be accelerated by neural networks, and at the same time, it can provide guidance and constraints for the learning process of neural networks through its rich structured information. This enhancement is reflected in the field of "symbol for neural", especially in two major application directions: KG-driven zero-shot and few-shot learning\cite{lee2018multi,wang2018zero,kampffmeyer2019rethinking,yao2020graph}, and knowledge-enhanced pre-trained language model (LM)\cite{liu2020k,zhang2019ernie}. These applications demonstrate how KG-based symbol grounding\footnote{Symbol grounding originates from cognitive science\cite{harnad1990symbol} and aims to connect abstract symbols to real-world physical entities. In the neural-symbolic field, it connects data to entities in a symbolic system, resulting in enhanced understanding and interpretability of the network.} can uniquely enhance the functionality of neural networks, making network models more robust and interpretable.

\subsubsection{KG-driven zero-shot and few-shot learning}

Knowledge graph-based Zero-Shot Learning (ZSL) models\cite{larochelle2008zero, socher2013zero, xian2016latent} perform well in addressing application problems in the visual field, especially in image recognition and classification. These models can recognize categories that were unseen during training by introducing additional knowledge (e.g., relationships between entities), which enhances the generalization ability of the model.

The \textbf{Zero-shot Learning via Semantic Embeddings and Knowledge Graphs(SEKG-ZSL)}\cite{wang2018zero}, a cutting-edge zero-shot learning framework for image classification, skillfully integrates semantic embeddings with KG, using them as inputs to GCN to train classifiers that recognize unseen categories. In this model, we utilize pre-trained text models (e.g., GloVe\cite{pennington2014glove} or word2vec\cite{mikolov2013efficient}) to extract semantic embedding vector representations of different categories in a high-dimensional space that capture the semantic properties of the corresponding categories. Meanwhile, the KG depicts the associations between categories graphically to aid model comprehension. In the graph structure, nodes represent different categories, and edges reveal the semantic relationships between categories. Then, the model processes these two types of knowledge inputs through the GCN to transfer and combine the information between different categories with the help of the associative relationships in the graph, thus developing a classifier that can generalize to unseen categories. In particular, it is noted that the KG sets constraints for the network's learning by utilizing the relationships between categories to guide the information transfer when training the classifier, and enhances its ability to learn and predict the unseen categories. Simply speaking, the KG plays the role of a map in the model, guiding the GCN to learn and infer along the correct semantic path. Overall, the SEKG-ZSL framework can effectively combine the semantic and visual information of categories to improve the accuracy and generalization ability of zero-sample learning.

\begin{figure}[htb] 
\centering 
\includegraphics[width=1\textwidth]{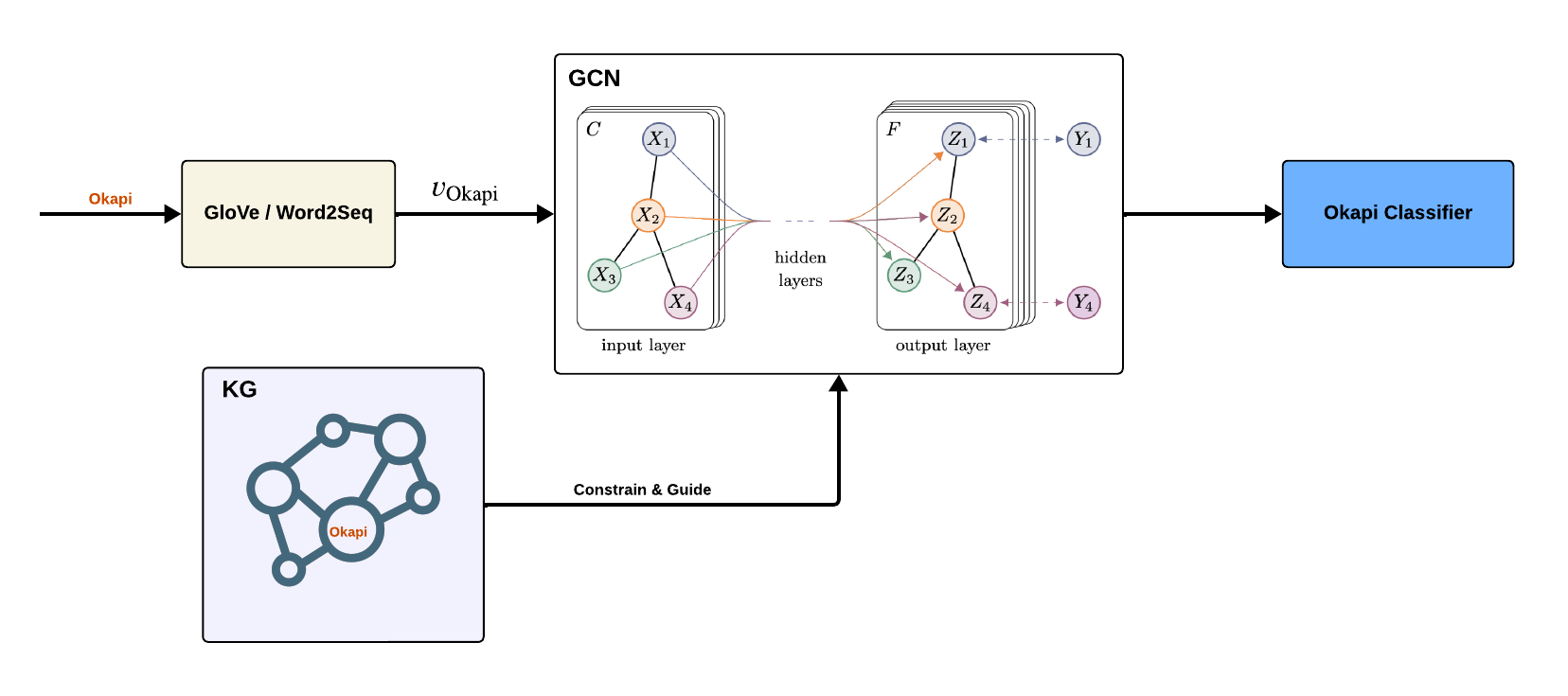} 
\caption{The basic architecture of training classifier by GCN with KG and semantic embedding} 
\label{figure 9}
\end{figure}

Typically, SEKG-ZSL performs well in dealing with zero-sample learning with single instance single label, but real-world scenarios such as image annotation usually involve a single image with multiple labels\cite{nam2014large, wei2014cnn, wang2016cnn,yeh2017learning}. For this reason, the \textbf{Multi-label Zero-shot Learning(ML-ZSL)}\cite{lee2018multi} method proposed by Lee C.W. et al. was developed to adapt to such problems. ML-ZSL achieves accurate prediction of unseen category labels by integrating structured KGs and effectively capturing the correlations among different labels. Its core lies in the information propagation mechanism, which is based on two key aspects: first, we need to construct a structured KG with nodes representing individual labels and update the node states through gated recurrent units (GRUs)\cite{cho2014learning} to simulate the information interactions and influences among labels and facilitate the propagation of information in the structured graph.
Also, the updating process can be encapsulated by the following equation \ref{eq 6}:
\begin{equation}
\label{eq 6}
    h_v^{(t)} = \text{GRU Cell}(u_v^{(t)}, h_v^{(t-1)})
\end{equation}
This equation updates the state \( h_v^{(t)} \) of node \( v \) at time step \( t \) by combining the information \( u_v^{(t)} \) from neighboring nodes and the node's previous state \( h_v^{(t-1)} \) via a GRU cell. This mechanism facilitates the propagation and updating of information throughout the graph. Second, the propagation matrix is learned to provide a strategy for the flow of information through the graph. This process determines the propagation weights through a relational function, where the weights are set based on the semantic word embedding of the labels. This approach ensures that the model can make effective inferences based on learned semantic relations even when faced with unseen labels. In summary, by combining structured KG propagation and propagation matrix learning, ML-ZSL not only finely regulates the information flow (Specific representations of knowledge-guided neural networks), but also significantly improves the prediction accuracy of unseen labels while maintaining the correct relationships between labels.

Methods based on GCN for zero-shot learning have shown great potential. However, a major challenge for these approaches is the knowledge dilution caused by excessive Laplacian smoothing\cite{smith1988edge} in multilayer GCN architectures, which makes the feature representations too similar and thus reduces the model's prediction performance for new categories. To address this problem, M Kampffmeyer et al. introduced the \textbf{Dense Graph Propagation (DGP)}\cite{kampffmeyer2019rethinking} method, which effectively avoids excessive information smoothing by establishing direct connections between nodes that are far away from each other in the KG, while optimizing information propagation by using distance-based weights between nodes. In specific terms, the module consists of a two-stage training procedure: the DGP is first trained to predict the weights of the last layer of the CNN for known categories, and then these predicted weights are applied to the CNN and fine-tuned to fit the new classifier. This approach not only enhances the prediction ability for unseen categories but also maintains the strong performance for known categories.

In addition to the zero-shot learning we previously discussed, few-shot learning also focuses on how to enable models to learn and generalize quickly in scenarios with scarce data. Within this framework, the \textbf{Graph Few-shot Learning(GFL)}\cite{yao2020graph} model employs its unique strategies to implement knowledge transfer, which migrates knowledge from the auxiliary graph to the target graph\cite{shervashidze2011weisfeiler, koutra2013deltacon}, thus helping to improve semi-supervised node classification in graphs. To be specific, GFL focuses on two types of relational structures at the node level (local perspective) and the graph level (global perspective) to facilitate knowledge transfer. First, at the node level, we aim to learn the associative structure among nodes of the same class. For this purpose, we introduce the Prototype Graph Neural Network (PGNN)\footnote{Prototype Graph Neural Network (PGNN)\cite{zhang2022protgnn} combines GNNs with prototype learning. By learning representative instances (prototypes) of key structures in a graph, GNN can more closely integrate the representation of the graph data with the representation of the associated prototypes, thus enhancing the interpretability of the representation in downstream tasks.}\cite{zhang2022protgnn}, which constructs the relational structure \( R_k \) from the sample set \( S_k \) of each class \( k \) to learn the prototype representation \( c_k \) of each class, thereby capturing the global relationships among nodes within the same category. And, this relational structure is typically based on metrics like k-hop common neighbors or topological distances. Also, the principle of prototype computation can be understood through the following equation \ref{eq 7}:
\begin{equation}
\label{eq 7}
    c_k = \text{Pool}\left( \text{PGNN}_\phi(R_k, f_{\theta}(S_k)) \right) 
\end{equation}
Here, \( \text{Pool} \) is a pooling operation (such as max pooling or average pooling) used to obtain a single prototype vector from the output of PGNN, \( f_{\theta}(S_k) \) represents the feature representation of all nodes in category \( k \) processed through a function with parameters \( \theta \), and \( \text{PGNN}_\phi \) denotes the PGNN model characterized by the parameter \( \phi \). Secondly, at the graph level, our goal is to learn the global information of a specific graph through the Hierarchical Graph Representation Gate and utilize this information to enhance the learning effectiveness of the network at the node level. This involves capturing the structural features of the graph at different levels and aggregating these features to form a comprehensive representation of the graph, and then adjusting the parameters \( \phi \) of the graph neural network using a gating mechanism, enabling the model to capture and utilize global structural information at the graph level.

\subsubsection{Knowledge-enhanced pre-trained language models}
Pre-trained language models(PLMs) based on knowledge enhancement improve the understanding and generation of text by incorporating external knowledge, such as KG. These integration methods utilize the structured information in KG to enable the language model to not only process complex text more logically, but also to accurately reference and reason about specialized knowledge, thus making the output more reliable.

In the field of Natural Language Processing(NLP), although the Bidirectional Encoder Representations from Transformers model (BERT)\cite{devlin2018bert} performs well on many tasks, there are still challenges when dealing with complex tasks that require deep knowledge and understanding. For this reason, researchers have explored the integration of structured KGs into BERT, aiming to provide the necessary external knowledge to enhance its comprehension and reasoning capabilities. \textbf{Knowledge-enabled BERT(K-BERT)}\cite{liu2020k} and \textbf{knowledge enhanced BERT(KnowBERT)}\cite{peters2019knowledge} are two advanced models for integrating KGs to enhance BERT, which incorporates enriched knowledge into BERT through different approaches, with the same goal but different core implementation strategies.

\textbf{KnowBERT}\cite{peters2019knowledge} introduces an innovative mechanism known as Knowledge Attention and Recontextualization (KAR), which adeptly bridges the gap between textual content and external knowledge bases. The essence of the KAR mechanism lies in its ability to identify entity mentions within the text and link them to corresponding entities in the knowledge base, thereby enriching the original contextual representation of the text. This process encompasses several critical steps: firstly, mention-span representation transforms potential entity mentions within the text into vectors of uniform dimensionality; secondly, an entity linker is employed to accurately align each mention with the most relevant entity in the knowledge base (symbol for neural); thirdly, by incorporating mention-span self-attention mechanism and through knowledge enhancement and recontextualization\cite{vaswani2017attention} steps, the selected entity information is effectively reintegrated into the textual representation, achieving a comprehensive integration of textual meaning and knowledge content. The basic architecture of KAR is shown in the following.

\begin{figure}[htb] 
\centering 
\includegraphics[width=0.9\textwidth]{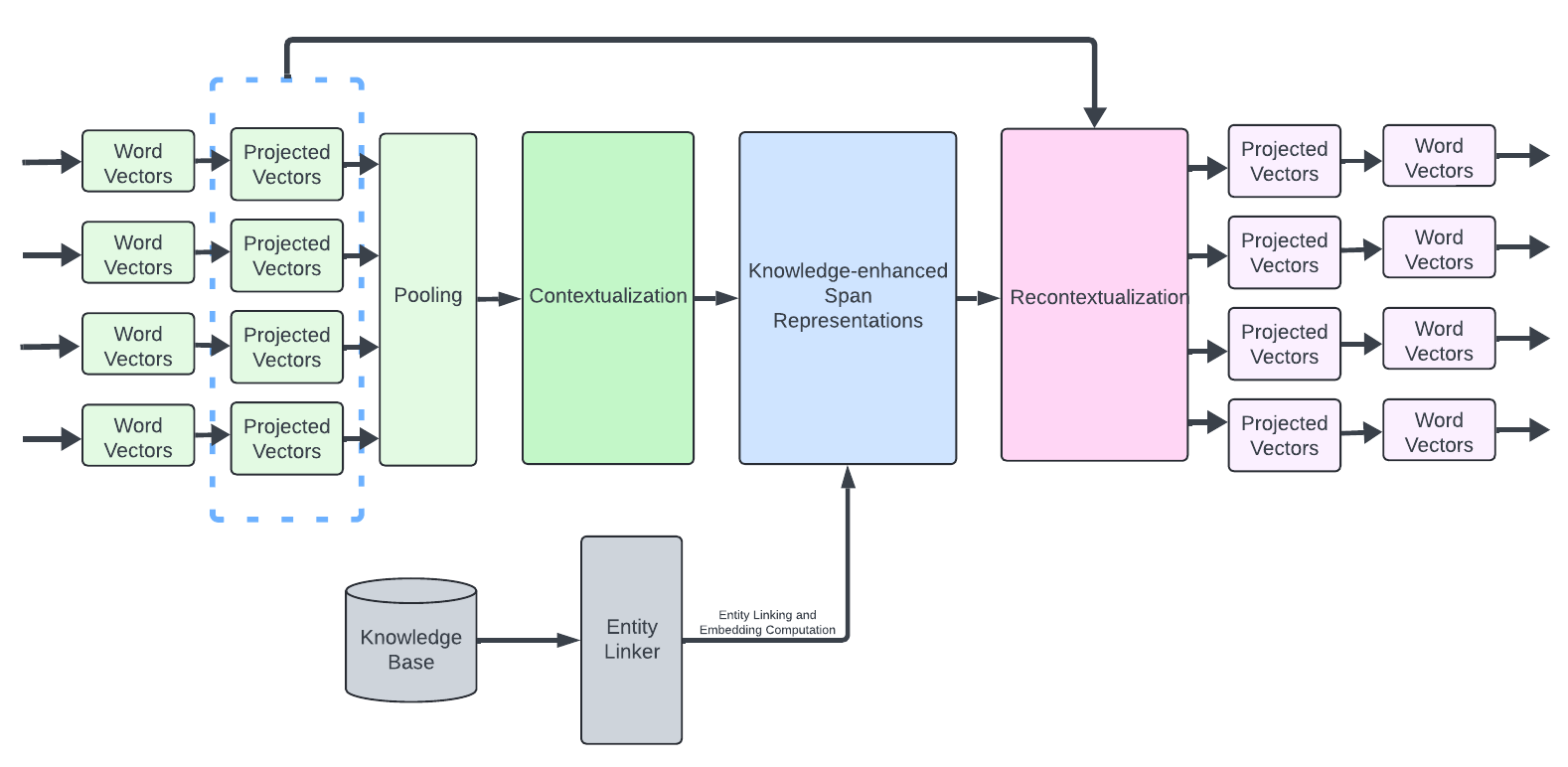} 
\caption{The Knowledge Attention and Recontextualization (KAR) component} 
\label{figure 10}
\end{figure}

On the other hand, the proposed \textbf{K-BERT}\cite{liu2020k} aims to solve the two main problems faced by language representation models based on knowledge injection: Heterogeneous Embedding Space (HES) and Knowledge Noise (KN). Firstly, to deal with the HES problem, K-BERT constructs a knowledge-rich sentence tree by directly integrating the KG information into the text, which realizes a unified representation of text and KG information in the same vector space, and effectively mitigates the HES problem. Secondly, to cope with the KN problem, K-BERT introduces soft-position embedding and a visible matrix mechanism. Soft positional embedding enables the model to integrate the injected knowledge while maintaining the text order without destroying the original text structure. Then, the visibility matrix controls the "visibility" of words to ensure that only relevant knowledge is computed, reducing the disturbance of irrelevant knowledge and mitigating the KN problem effectively. The visible matrix \(M\) is defined by the equation \ref{eq 8}:
\begin{equation}
\label{eq 8}
    M_{ij} = \begin{cases} 0 & \text{if } w_i \leftrightarrow w_j \\ -\infty & \text{otherwise} \end{cases} 
\end{equation}
Here, \(w_i\) and \(w_j\) represent tokens within the sentence, and \(w_i \leftrightarrow w_j\) denotes that \(w_i\) and \(w_j\) are within the same branch of the knowledge-rich sentence tree and therefore can "see" each other. If they are not in the same branch, the value of \(-\infty\) could effectively mask out (i.e., makes invisible) the token \(w_j\) when computing attention for \(w_i\), ensuring that only contextually relevant knowledge influences the representation of each token.

\textbf{Conclusion:} In the paradigm of neural for symbol, KG has provided neural network learning with a wealth of structured knowledge and deep semantic and contextual information, serving as important guidance during the learning process. Faced with the challenge of scarce samples, the structured information from KGs significantly enhances the neural network's learning capabilities in complex language tasks. For instance, models such as SEKG-ZSL\cite{wang2018zero} and GFL\cite{yao2020graph} effectively learn through semantic embeddings and graph structure reasoning, even in situations with extremely limited samples. Moreover, in handling natural language processing (NLP) tasks, KGs provide the models with rich contextual information and precise factual knowledge that are difficult to obtain from traditional pre-training. This information directly enhances the models' understanding of complex queries and improves their ability to handle terms and concepts specific to certain fields. Additionally, it ensures the factual and logical consistency of the generated text, thereby enhancing the model's interpretability.

\subsection{Hybrid neural-symbolic integration}
Compared to the two previously mentioned categories, our hybrid integration model focuses more on processing the functions of neural networks and symbolic systems in parallel, allowing them to operate independently without interfering with each other. Through specific mechanisms, these two systems can share information and results. Typically, we expect this parallel operation to promote and grow with each other, forming a cyclic enhancement learning model: the output of each system can become part of the input of the other system, thus driving the iterative progress of the whole system. In addition, this learning model is also expected to apply to a wide range of application tasks, including Q\&A systems\cite{ding2019cognitive}, KG-to-text\cite{ke2021jointgt}, entity alignment\cite{zhang2023iterative} and sentiment identification\cite{zhao2023knowledge}, etc.

\begin{figure}[htb] 
\centering 
\includegraphics[width=0.7\textwidth]{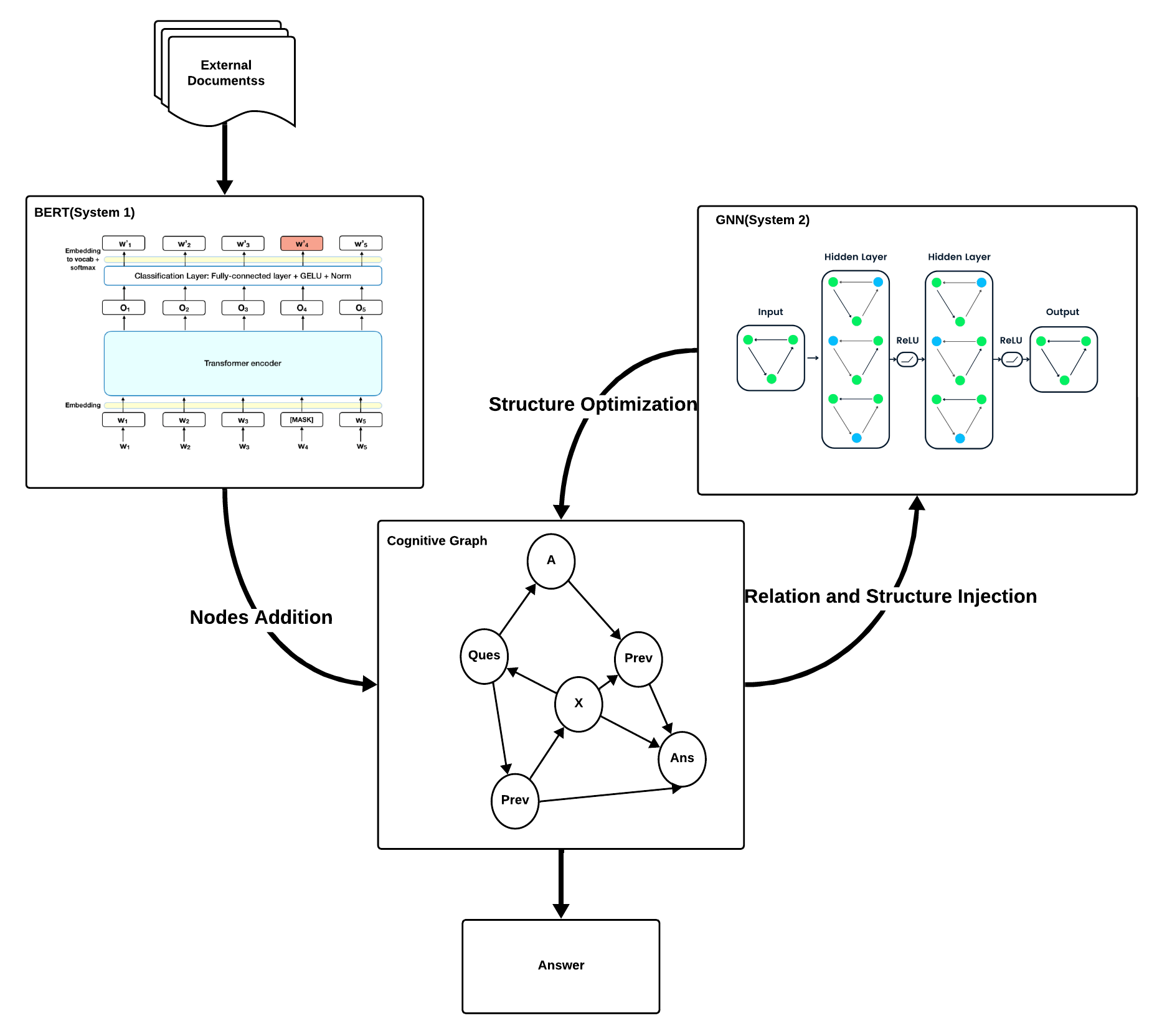} 
\caption{Overview of CogQA implementation} 
\label{figure 11}
\end{figure}

In Section \ref{Section 1}, we discussed the concepts of System 1 and System 2\cite{kahneman2011thinking} from cognitive science. Based on these ideas, Ding M and his team combined the capabilities of BERT (acting as System 1) and GNN (acting as System 2) to develop the \textbf{Cognitive Graph QA(CogQA)}\cite{ding2019cognitive} model. This model employs a cognitive graph(KG-like structure) to mimic human dual-process cognition, aiming to address the complex challenges encountered when performing multi-hop question answering across large-scale document sets. CogQA constructs a dynamic cognitive graph that links information dispersed across multiple documents and iteratively mines and verifies potential answers. Specifically, apart from initializing the graph (creating starting nodes based on entities mentioned in the questions and marking them as "frontier nodes") and determining the termination conditions, each iteration of building the cognitive graph primarily involves three steps: First, relevant information, including potential answers and related entities, is extracted from the document set using BERT\cite{devlin2018bert} (system 1); Second, based on the information provided by system 1, CogQA updates the cognitive graph by adding new nodes and edges, where new nodes include entities or answer candidates identified from the text, and edges represent logical relationships between entities. These new nodes are also marked as frontier nodes for use in the next iteration; Third, once the cognitive graph is updated, GNN(system 2) begins analyzing the logic and relationships between entities in depth and optimizing the structure of the cognitive graph. Importantly, after each iteration, CogQA assesses whether the cognitive graph is sufficiently comprehensive to answer the original question. If the information is deemed adequate, the model will predict the final answer based on the current cognitive graph. Otherwise, the model will continue iterating, further expanding, and refining the cognitive graph with each cycle until a satisfactory answer is found. The basic architecture and implementation of CogQA is shown in Figure \ref{figure 11}

In the task of Knowledge Graph to Text Generation (KG-to-text)\cite{gardent2017webnlg}, existing models often ignore the structural information of the graph or lack pre-training tasks for accurately modeling graph-text alignment. To address these issues, a joint graph-text representation learning model called \textbf{JointGT}\cite{ke2021jointgt} is proposed. This model preserves the structure of the input graph by introducing structure-aware semantic aggregation modules in each Transformer layer of the encoder. In addition, the model designs three new pre-training tasks to explicitly enhance graph-text alignment, including graph-enhanced text reconstruction, text-enhanced graph reconstruction, and graph-text embedding alignment via optimal transport. It is worth noting that these pre-training tasks are performed iteratively to continuously optimize the performance of the model. In the graph-text embedding alignment task, it employs the optimal transport theory to minimize the cost between graph and text embedding, which is a typical iterative process. In this process, the model calculates the distance between the embedding vectors of the graph and the text and adjusts these embedding vectors by the optimal transport algorithm so that the transit cost is minimized. In this way, the model is better able to capture and learn the complex relationships between graphs and texts, thus generating more accurate and coherent texts.

Entity alignment is an important task in the field of KGs, aiming at identifying entities with the same or similar meanings in different KGs. Traditional approaches usually rely on extensive human involvement or simple embedding techniques, which may not be sufficient to capture the rich relationship and attribute information among entities. However, the \textbf{HGNN-EA}\cite{zhang2023iterative} model significantly improves the accuracy of entity alignment by employing Heterogeneous Graph Neural Networks (HGNNs)\cite{zhang2019heterogeneous} to process entity and relationship data in the KG and using iterative fusion methods to enhance model training. In this model, different types of nodes (entities and relations) are modeled simultaneously and their semantic representations are strengthened through dynamic interactions during iterations. Each iteration optimizes the information exchange between nodes through Graph Attention Network (GAT)\cite{velivckovic2017graph} to represent their semantic and structural associations more comprehensively. In addition, HGNN-EA applies a distance-based approach in entity alignment, which can accurately measure the semantic distances between entities in different KGs, and effectively identify pairs of entities with the same or similar meanings.

Implicit sentiment recognition, a common task in text analysis of the NLP field, faces challenges due to the lack of structural information and noise in predefined graph structures. To address these challenges, the \textbf{Knowledge-Fusion-Based Iterative Graph Structure Learning Framework (KIG)}\cite{zhao2023knowledge} has been introduced. This innovative method aims to construct rich initial graph structures by integrating knowledge from multiple perspectives—including co-occurrence statistics, cosine similarity, and syntactic dependency trees—to more comprehensively capture the subtle expressions of sentiment within texts. The core mechanism of KIG lies in its iterative evolutionary graph learning process. In each iteration, KIG evaluates the effectiveness of the current graph structure and updates it based on the information learned, thereby achieving optimal node representations and graph topology. During the iterative process, KIG employs a multi-view fusion strategy, integrating graph structural perspectives from different information sources to form a comprehensive and expressive graph representation. For example, it considers both semantic similarity and syntactic structure to capture the textual sentiment tendencies more thoroughly. Moreover, each iteration involves refining the graph structure by adjusting the adjacency matrix and utilizing GCNs\cite{{schlichtkrull2018modeling,kipf2016semi}} to extract and merge node features, subsequently updating the node embedding.

\textbf{Conclusion:} The four described models of Hybrid neural-symbolic integration, although applied in different domains, share a key feature — iterative learning mechanisms. Iterative learning significantly enhances the models' reasoning capabilities and data representation through a continuous cyclic process. For instance, in the HGNN-EA\cite{zhang2023iterative} and CogQA\cite{ding2019cognitive}, iterative methods are used to progressively optimize the representation of nodes within KGs, thereby more accurately reflecting the complex relationships and attributes between entities. This iterative process not only improves the models' understanding of data structures but also refines the application of symbolic logic with each iteration, thereby increasing the accuracy and efficiency of problem-solving. Iterative learning also enhances the models' adaptability to new situations and their explanatory power, making them more flexible in practical applications. Through continual iterative updates, the models accumulate learning experiences, optimize their decision-making processes, and exhibit greater robustness and adaptability when confronted with unknown data or complex scenarios.

\section{Future Trends \& Direction}
\label{Section4}
\subsection{Multimodal and multidomain learning}
Multimodal and multidomain learning represents a significant trend in the field of deep learning\cite{kannan2020multimodal, chen2022hybrid, mousselly2018multimodal, sun2020multi, zhu2022multi}, offering unprecedented opportunities but also presenting considerable challenges, particularly in terms of information fusion and domain adaptation. The challenge of information fusion primarily involves effectively integrating data from diverse modalities, such as text, images, and audio, which differ significantly in their forms and processing methods. By utilizing KG as a unified semantic framework, we can better align and integrate data from these varied modalities. The structured information within the KG provides essential context, facilitating the fusion of information from different sources, and thereby enhancing the overall understanding capabilities of the model. On the other hand, the issue of domain adaptation focuses on enabling models to operate across different domains while maintaining performance amidst variations in data distribution and characteristics across these domains. Employing a KG as a bridge for cross-domain data processing, and incorporating common knowledge and rules across domains, can significantly enhance the adaptability and generalization capabilities of models in new scenarios.

\subsection{Reasoning efficiency}
Improving reasoning efficiency is one of the challenges in achieving deep learning productization, especially in scenarios that require fast responses, such as mobile device applications, self-driving, and other real-time processing systems. These application scenarios typically require models to perform tasks quickly and accurately with limited computational resources. KGs may play a key role in this environment by providing neural network models with rich prior knowledge to optimize reasoning paths. More specifically, KGs can predefine the logical relationships and rules required during the inference process, allowing the model to directly perform some of the decisions and computations without the need for deep neural computation. The rules and known facts in the KG can be utilized for direct processing when performing reasoning tasks, thus reducing the reliance on data-driven reasoning. This approach reduces the amount of computation and also increases the speed of reasoning, making the model more efficient and practical in real applications.
\subsection{Graph-integrated Transformer}
Although graph neural networks (GNNs)\cite{scarselli2008graph} have demonstrated excellent capabilities in processing structured data, especially KGs, in practice, GNNs face challenges such as low computational efficiency, limited scalability, and lack of performance when dealing with large-scale datasets. The computational complexity of GNNs stems mainly from the sparseness and irregularity of graph data, which makes the optimization and acceleration of these models particularly difficult. Given these challenges, it is particularly promising to focus future research and applications on combining KGs with Transformer-based models\cite{vaswani2017attention}, which, by their self-attentive mechanism, can efficiently process large-scale datasets and effectively capture long-distance dependencies, and especially excel in the processing of text and other continuous data streams. Specifically, by directly incorporating entities and relationships from the KG into the Transformer's self-attention mechanism, the knowledge-enhanced Transformer model can provide clear inference paths and logical proofs while maintaining efficient data processing.

\section{Conclusion}
\label{Section5}
In this paper, we explore how KG-based neural symbolic integration can be applied in three different categories. This research comprehensively demonstrates the potential of combining deep learning with KG reasoning, providing a theoretical foundation for the future development of more interpretable and efficient AI systems. The goal of this comprehensive approach is to bridge the gap between intuitive human reasoning and machine execution, thereby enhancing the utility and transparency of AI applications in multiple domains.

\begin{appendices}

\section{Supplemental Material} 
\label{appendix:supplemental} 

\subsection{Knowledge Graph} 
\label{appendix:Knowledge Graph}
A knowledge graph(KG) is a structured representation of knowledge where entities, their attributes, and the relationships between them are organized into a graph-like structure. In mathematical terms, a KG can be represented as G = (V, E), where V represents the set of vertices or nodes corresponding to entities, and E represents the set of edges or relationships between these entities. Each relationship is often expressed as a triplet (subject, predicate, object), where the subject and object are entities and the predicate represents the relationship between them. Mathematically, this triplet can be denoted as (s, p, o). Each entity is associated with a set of properties or attributes, forming a vector space representation. Mathematically, this can be denoted as \( V_i = \{p_1, p_2, ..., p_n\} \), where \( V_i \) represents the ith entity and \( p_1, p_2, ..., p_n \) represent its attributes. Relationships between entities are represented as directed edges connecting nodes in the graph. By leveraging graph theory and mathematical models, KGs enable the organization, retrieval, and analysis of complex information in a structured and interconnected manner, facilitating various applications such as semantic search, question answering, and knowledge discovery.

\subsection{Graph neural networks}
Graph Neural Networks (GNNs)\cite{scarselli2008graph} belong to a special class of neural networks that are designed to operate on data following the structure of a graph, in which there are entities or nodes, and relations or edges between them. That's pretty uncommon, actually. Unlike conventional neural networks, which can learn and reason from grid-like data, GNNs are quite able to get pattern recognition within data that's highly irregular and interconnected. GNNs iteratively update the representations of the nodes by allowing them to pool information from neighboring nodes, such that they can represent complex dependencies and patterns of the graph structure. This is the process, usually message passing between nodes, whereby each node aggregates information from its neighbors and changes its own representation based on the information.
\subsubsection{Graph convolutional networks}
Graph convolutional networks (GCNs) \cite{kipf2016semi}are, in fact, convolutional neural networks (CNNs) that generalize the classical CNNs' convolutional operation from regular data domains to irregular ones, such as graphs. Being directly performed over the graph structure, convolutional operations in GCNs allow for the capturing of localized and global information in a principled manner over the corresponding adjacent nodes. Central to the idea of GCNs is the aggregation of feature information from the node's neighborhood, normally understood to mean the immediate neighbors, followed by updating the node representation with the gathered representation, and repeating this process several times over multiple layers. This is being done several times over a few layers to progressively learn more complex hierarchical representations of the graph data.
\subsubsection{Graph attention networks}
Graph Attention Networks (GATs)\cite{velickovic2017graph} are another type of graph neural network that applies attention mechanisms to allow the model to capture complex dependencies between different nodes in graph-structured data, within natural language processing. In traditional graph convolutional networks (GCNs), information from neighbors of nodes is uniformly aggregated, but the attention of GAT is dynamically computed for each pair of neighbor nodes and central node, which enabled our model to generalize the convolutional model by focusing more on the relevant neighbors of each node. This attention mechanism, in other words, allows the GATs to effectively learn the importance in light of their relevance to the target node, capturing differences in level that arise from the graph structure.
\end{appendices} 


\end{document}